\documentclass[11pt,a4paper]{article}
\usepackage[utf8]{inputenc}
\usepackage[hyperref]{acl2019}
\usepackage{times}
\usepackage{latexsym}

\usepackage{url}
\usepackage{amsmath}
\usepackage{amssymb}
\usepackage{graphicx}
\usepackage{multirow}
\usepackage{array}
\usepackage{arydshln}
\usepackage{enumitem}
\usepackage{float}
\usepackage{microtype}
\usepackage[absolute]{textpos}

\aclfinalcopy 
\def\aclpaperid{69} 

\usepackage[normalem]{ulem}
\definecolor{darkgreen}{rgb}{0.0, 0.5, 0.0}

\def\VRdel#1{\bgroup\markoverwith{\textcolor{magenta}{\rule[0.5ex]{2pt}{1pt}}}\ULon{#1}}

\def\ODdel#1{\bgroup\markoverwith{\textcolor{darkgreen}{\rule[0.5ex]{2pt}{1pt}}}\ULon{#1}}

\def\YKdel#1{\bgroup\markoverwith{\textcolor{blue}{\rule[0.5ex]{2pt}{1pt}}}\ULon{#1}}

\def\KSdel#1{\bgroup\markoverwith{\textcolor{orange}{\rule[0.5ex]{2pt}{1pt}}}\ULon{#1}}

\newcommand\Bstrut{\rule[-0.9ex]{0pt}{0pt}} 

\title{Automatic Quality Estimation for Natural Language Generation: \\ Ranting (Jointly Rating and Ranking)}

\author{Ondřej Dušek,$^\ast$ Karin Sevegnani,$^\dagger$ Ioannis Konstas$^\dagger$ \and Verena Rieser$^\dagger$ \\
  $^\ast$Charles University, Faculty of Mathematics and Physics, Prague, Czechia \\
  $^\dagger$Heriot-Watt University, MACS, The Interaction Lab, Edinburgh, Scotland, UK  \\
  \texttt{odusek@ufal.mff.cuni.cz, \{ks85,i.konstas,v.t.rieser\}@hw.ac.uk} \\}

\date{}

\begin{document}
\maketitle

\begin{textblock*}{\textwidth}(2.5cm,1cm)
In \emph{Proceedings of INLG}, Tokyo, Japan, October 2019.
\end{textblock*}

\begin{abstract}
We present a recurrent neural network based system for automatic quality estimation of natural language generation (NLG) outputs, which jointly learns to assign numerical \emph{ratings} to individual outputs and to provide pairwise \emph{rankings} of two different outputs. The latter is trained using pairwise hinge loss over scores from two copies of the rating network.

We use learning to rank and synthetic data to improve the quality of ratings assigned by our system: we synthesise training pairs of distorted system outputs and train the system to rank the less distorted one higher. This leads to a 12\% increase in correlation with human ratings over the previous benchmark.
We also establish the state of the art on the dataset of relative rankings from the E2E NLG Challenge \cite{dusek_evaluating_2019}, where synthetic data lead to a 4\% accuracy increase over the base model.
\end{abstract}

\section{Introduction}\label{sec:intro}
%
%

While 
automatic output quality estimation (QE) is an established field of research in other areas of NLP, such as machine translation (MT) \cite{specia_machine_2010,specia_quality_2018}, 
research on QE in natural language generation (NLG) from structured meaning representations (MR) such as dialogue acts is relatively recent 
\cite{dusek_referenceless_2017,ueffing_quality_2018} and often focuses on output fluency only \cite{tian_treat_2018,kann_sentence-level_2018}.
In contrast to traditional metrics, QE does not rely on gold-standard human reference texts \cite{specia_machine_2010}, which are expensive to obtain, do not cover the full output space, and are not accurate on the level of individual outputs \cite{novikova_why_2017,reiter_structured_2018}. 
Automatic
QE for NLG has 
several possible use cases that can improve NLG quality and reliability. For example, rating individual NLG outputs allows to ensure a minimum output quality and engage a backup, e.g., template-based NLG system, if a certain threshold is not met. Relative ranking of multiple NLG outputs can be used directly within a system to rerank n-best outputs or to guide system development, selecting optimal system parameters or comparing to state of the art.

In this paper, we present a novel model that jointly learns to perform both tasks---rating individual outputs as well as pairwise ranking. We show that this leads to performance improvements over previously published results \cite{dusek_referenceless_2017}. Our model is portable, since we do not assume any specific input schema and only rely on ratings of the text output, which are relatively easy to obtain, e.g.\ through crowdsourcing for a small number of outputs of an initial NLG system.
The model learns to rank or rate according to any criterion annotated in the data, such as adequacy, fluency, or overall quality \cite[see~e.g.,][]{wen_semantically_2015,manishina_automatic_2016,novikova_why_2017}.
Our main contributions are as follows:
\begin{itemize}[itemsep=0pt,topsep=1pt,leftmargin=10pt]
\item A novel, domain- and input representation-agnostic, and conceptually simple model for NLG QE, which jointly learns ratings and pairwise rankings. It is able to seamlessly switch between the two and is directly applicable for n-way ranking (see Section~\ref{sec:model}).
Crucially, it does not require human-authored references during inference.

\item An original methodology for synthetically generating training instances for pairwise ranking based on introducing errors (see Section~\ref{sec:synth-data}).
\item A significant, 12\% relative improvement in Pearson correlation with human ratings over results previously published on the dataset of \citet{novikova_why_2017}, as well as the first pairwise ranking results for NLG QE on the E2E ranking dataset of \citet{dusek_evaluating_2019}, with significant improvements over the baseline due to synthetic training instance generation (see Sections~\ref{sec:experiments} and~\ref{sec:results}).
\end{itemize}
Both datasets are freely available, and we release our experimental code on GitHub.\footnote{The datasets can be downloaded under the following links: \url{https://github.com/jeknov/EMNLP_17_submission}, \url{http://www.macs.hw.ac.uk/InteractionLab/E2E/}. Our code is available at \url{https://github.com/tuetschek/ratpred}.}

\section{The Task(s)}\label{sec:task}

\begin{figure*}[tb]

\centering\small
\setlength{\extrarowheight}{2pt}
\begin{tabular}{>{\raggedright\arraybackslash}m{2.5cm}>{\raggedright\arraybackslash}m{10cm}>{\centering\arraybackslash}m{2cm}}\hline
\multicolumn{2}{c}{\bf Instance} & \bf Rating/Rank \\\hline
\bf MR & inform\_only\_match(name=`hotel drisco', area=`pacific heights') & \multirow{2}{*}{4}\\
\bf \textsc{RNNLG} output & the only match i have for you is the hotel drisco in the pacific heights area.\\\hline
\bf MR & inform(name=`The Cricketers', eat\_type=`coffee shop', rating=high, family\_friendly=yes, near=`Café Sicilia') &  \\
\bf \textsc{Zhang} output &  The Cricketers is a children friendly coffee shop near Café Sicilia with a high customer rating . & \emph{better} \\
\bf \textsc{TR2} output & The Cricketers can be found near the Café Sicilia. Customers give this coffee shop a high rating. It's family friendly. & \emph{worse} \\\hline
\end{tabular}

\caption{Examples for NLG output quality rating (top, from the NEM dataset) and ranking (bottom, from the E2E rankings dataset); \textsc{RNNLG}, \textsc{Zhang} and \textsc{TR2} are NLG systems. See Section~\ref{sec:data} for details on the datasets.}
\label{fig:examples}
\end{figure*}

The task of NLG QE for \emph{ratings} is to assign a numerical score to a single NLG output, given its input MR, such as a dialogue act (consisting of the main intent, attributes and values). The score can be e.g.\ on a Likert scale in the 1-6 range \cite{novikova_why_2017}.
In a pairwise \emph{ranking} task, the QE system is given two outputs of different NLG systems for the same MR, and decides which one has better quality (see Figure~\ref{fig:examples}).

As opposed to automatic word-overlap-based metrics, such as BLEU \cite{papineni_bleu:_2002} or METEOR \cite{lavie_meteor:_2007}, no human reference texts for the given MR are required. This widens the scope of possible applications -- QE systems can be used 
for previously unseen MRs.

%
%
\section{Model}\label{sec:model}

\begin{figure}[tb]
\centering
\includegraphics[width=\columnwidth]{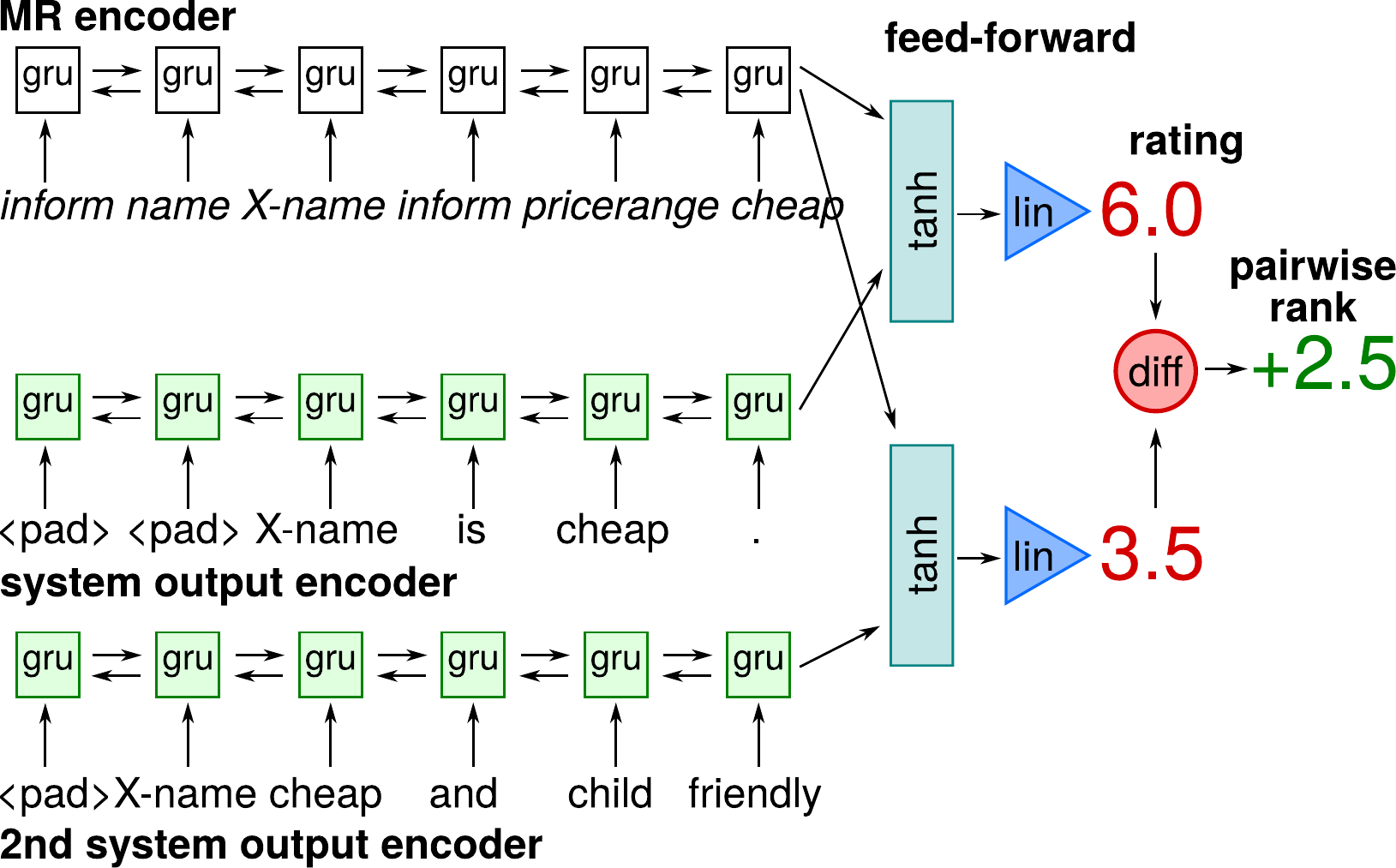}
\caption{Schematic of our NLG QE model. Components sharing weights are shaded with the same colour.}
\label{fig:model}
\end{figure}

Our model is a direct extension of the freely available RatPred system \cite{dusek_referenceless_2017}. The original RatPred model assigns numerical ratings to single outputs and is a dual-encoder \cite{lu_practical_2017}, consisting of two GRU-based recurrent neural networks \cite{cho_learning_2014} encoding the MR and the system output, followed by fully connected layers and a final linear layer providing the score. The system is trained using squared error loss, and it uses dropout over embeddings \cite{hinton_improving_2012}.

We 
make RatPred's encoders bidirectional and add a novel extension to allow pairwise ranking---a second copy of the system output encoder plus the fully connected layers and linear layer (Figure~\ref{fig:model}). All network parameters are shared among the two copies. This way, the network is able to rate two NLG outputs at once. We add a simple difference operator on top of this; the pairwise rank is computed as the difference between the two predicted scores.
In addition to the squared loss for rating, we incur pairwise hinge loss for ranking. The final loss function looks as follows:
\begin{equation*}
\mathcal{L} = (1 - I)\cdot(\hat{y}–y)^2 + I\cdot \max(0, 1 - (\hat{y} - \hat{y}'))
\end{equation*}
$I$ indicates if the current instance is a ranking-based one (value of 1 for ranking and 0 for rating, effectively a mask to only incur the correct loss). $y$ denotes the true score for a NLG output, $\hat{y}$ and $\hat{y}'$ denote scores assigned by the model for (up to) two NLG outputs.\footnote{Since the ranking result is a difference of two scores assigned by copies of the same network, we can assume without loss of generality that $\hat{y}$ ranks higher than $\hat{y}'$.} Note that $\hat{y}'$ is ignored in rating instances, while a true score $y$ is ignored for ranking. This way, the same network performs ranking and rating jointly, and it can be exposed to training instances of both types in any order.
Our model is also directly applicable to \mbox{n-way} rankings---using it to score a group of NLG outputs and comparing the scores is equivalent to comparing the  pairwise ranking.

Jointly learning to rank and rate was first introduced by \citet{sculley_combined_2010} for support vector machines and similar approaches have been applied for image classification \cite{park_personalized_2017,liu_learning_2018} as well as audio classification \cite{lee_minimization_2016}, However, we argue that the application for text classification/QE is novel, as is the implementation as a single neural network with two parts that share parameters, capable of training from mixed ranking/rating instances with masking to incur the proper loss.

\section{Synthetic Training Data Generation}\label{sec:synth-data}

\begin{figure*}[tb]
\begin{center}
\small
\setlength{\extrarowheight}{2pt}
\begin{tabular}{ll}\hline
\bf MR &  inform(name='house of nanking',food=chinese)\\
\bf RNNLG output (0 errors) & house of nanking serves chinese food .\\\hdashline[0.5pt/2pt]
\bf 1 error & house of nanking \uline{restaurant} chinese food . \\
\bf 2 errors & house of nanking serves \uline{food} chinese food \uline{cheaply} . \\
\bf 3 errors & \uline{food} house of nanking \uline{house of nanking} serves chinese \uline{chinese} food .\Bstrut\\\hline
\end{tabular}
\caption{Synthetic data generation example.}\label{fig:synth-pairs}
\end{center}
\vspace{-1.5mm}
\fontsize{10pt}{12pt}\selectfont
\emph{Synthesising errors:} The original NLG output is distorted by introducing errors (underlined) of the following types: Words in the texts are removed or duplicated at original or random positions, random words from a dictionary learned from training data are used to replace current words or added at random positions. Other words are preferred over articles and punctuation in making the changes (see \citealp{dusek_referenceless_2017} for details).

\emph{Rating instances:} We use the same settings as \citet{dusek_referenceless_2017} for synthetic individual rating instances -- generating up to 4 errors and lowering the target rating by 1 each time (lowering by 2 if the original value was 6). We are able to generate more synthetic rating instances since \citet{dusek_referenceless_2017} did not use all available NLG system outputs due to a bug in their code (cf.\ Table~\ref{tab:results-nem}).

\emph{Ranking Instances:} Following our new method, pairs of outputs with a different number of errors (e.g., 0-1, 1-3) are then sampled as synthetic training instances for ranking. In our setting, we introduce up to 4 errors and create instances for all numbers of errors against the original (0-1 through 0-4), plus a set of 5 other, randomly chosen instances (e.g. 1-3, 2-4).
We use both rating and ranking synthetic instances for NEM data and only ranking synthetic instances for E2E data.
\end{figure*}

We use RatPred's code to generate synthetic rating instances from both NLG outputs and human-authored texts by distorting the text and lowering its score (i.e., randomly removing or adding words; cf. \citet{dusek_referenceless_2017} and Figure~\ref{fig:synth-pairs} for details).
We also create synthetic training pairs by using the same NLG output/human-authored text under two different levels of distortion (e.g., one vs.\ two artificially introduced errors). The system is then trained to rank higher the version of the text with fewer errors (see Figure~\ref{fig:synth-pairs} for an example).
This novel approach can be used to generate synthetic training data for both ranking and rating tasks---in a rating task, the generated ranking instances are simply mixed among the original training instances for rating, and the model uses both kinds for training.
Note that synthetic data are never used for validation or testing in any of our setups.

\section{Experimental Setup}\label{sec:experiments}

\subsection{Datasets}\label{sec:data}

We experiment on the following two datasets, both in the restaurant/hotel information domain:
\begin{itemize}[itemsep=0pt,topsep=1pt,leftmargin=10pt]
\item NEM\footnote{While the dataset authors did not give it a name, we use ``NEM'' as an acronym for ``New Evaluation Metrics'', which comes from the title of the paper.} \cite{novikova_why_2017} -- Likert-scale rated outputs (scores 1--6) of 3 NLG systems over 3 datasets, totalling 2,460 instances. 
\item E2E system rankings \cite{dusek_evaluating_2019} -- outputs of 21 systems on a single NLG dataset with 2,979 5-way relative rankings.
\end{itemize}
We choose these two datasets because they contain human-assessed outputs from a variety of NLG systems. 
Another candidate is
the WebNLG corpus \cite{gardent_webnlg_2017}, which we leave for future work due to MR format differences.

Although both selected datasets contain ratings for multiple criteria (informativeness, naturalness and quality for NEM and the latter two for E2E), we follow \citet{dusek_referenceless_2017} and focus on the overall quality criterion in our experiments as it takes both semantic accuracy and fluency into account.

We use RatPred's preprocessing, synthetic data generation, and 5-way cross-validation split on the NEM dataset. In addition, we generate synthetic training pairs as described in Section~\ref{sec:synth-data}.
We convert the 5-way rankings from the E2E set to pairwise rankings \cite{sakaguchi_efficient_2014} (leaving out ties), which produces 15,001 instances. We split the data into training, development and test sections in an 8:1:1 ratio, ensuring that each section contains NLG outputs for different MRs \cite{lampouras_imitation_2016}.\footnote{We first split the data according to MRs, then assign different MRs with all corresponding system outputs into different sections.}
In addition to the human-assessed NLG outputs themselves, human-authored training data for the NLG systems are also available and are used for synthetic instances.
We use a partial delexicalisation (replacing names with placeholders).\footnote{See Table~\ref{tab:delex} in the Supplementary for details.}

\subsection{Model Settings}\label{sec:setup}

We evaluate our model in several configurations, with increasing amounts of synthetic training data. Note that even setups using \emph{training} human references (i.e.\ additional in-domain data) are still ``referenceless''---they do not use human references for test MRs. Setups using human references for validation and \emph{test} MRs (``reference-aided''; marked with ``*'' in Table~\ref{tab:results-nem}) are not referenceless and are mainly shown for comparison with \citet{dusek_referenceless_2017}.

We use the same network parameters for all setups, selected based on a small-scale grid search on the development data of both sets, taking training speed into consideration.\footnote{See Table~\ref{tab:parameters} in the Supplementary for details.}
As a result, we use a network with fewer parameters than \citet{dusek_referenceless_2017}, which makes our base setup worse-performing than the original base setup, despite our use of bidirectional encoders (cf.~Section~\ref{sec:results}). On the other hand, training runs several times faster.
We use Adam \cite{kingma_adam:_2015} for training, evaluating on the validation set after each epoch and selecting the best-performing configuration. Synthetic data are removed after 50 (out of 100) epochs.
Following \citet{dusek_referenceless_2017}, we run all experiments with 5 different random initializations of the networks and report averaged results.

\subsection{Evaluation Metrics}\label{sec:metrics}

On the NEM data, we follow \citet{dusek_referenceless_2017} to compare with their results. We use Pearson correlation of system-provided ratings with human ratings as our primary evaluation metric; we also measure Spearman rank correlation, mean absolute error (MAE) and root mean squared error (RMSE).
On the E2E data, we use pairwise ranking accuracy (or Precision@1), a common ranking metric. We also measure mean ranking loss, i.e., mean score difference in wrongly rated instances.

\section{Results and Discussion}\label{sec:results}

\begin{table*}[tb]
\centering\small
\begin{tabular}{lrcccc}\hline
\bf System & \bf Training insts & \bf Pearson & \bf Spearman & \bf MAE & \bf RMSE \\\hline
Constant                                            &       - & -     & -     & 1.013 & 1.233\\
BLEU* \cite{papineni_bleu:_2002}                    &       - & 0.074 & 0.061 & 2.264 & 2.731\\
METEOR* \cite{lavie_meteor:_2007}                   &       - & 0.095 & 0.099 & 1.820 & 2.129\\
ROUGE-L* \cite{lin_rouge:_2004}                     &       - & 0.079 & 0.072 & 1.312 & 1.674\\
CIDEr* \cite{vedantam_cider:_2015}                  &       - & 0.061 & 0.058 & 2.606 & 2.935\\\hdashline[0.5pt/2pt]
RatPred \cite{dusek_referenceless_2017} base system &   1,476 & 0.273 & 0.260 & 0.948 & 1.258 \\
+ generated data based on training system outputs   &   3,937 & 0.283 & 0.268 & 0.948 & 1.273 \\
+ generated data based on training human references &  45,137 & 0.330 & 0.274 & 0.914 & 1.226 \\\hdashline[0.5pt/2pt]
+ generated data based on test human references*    &  80,522 & 0.354 & 0.287 & 0.909 & 1.208 \\\hline
Our base system                                     &   1,476 & 0.253 & 0.252 & 0.917 & 1.221 \\ 
+ generated data based on training system outputs   &   8,856 & \bf 0.332 & \bf 0.308 & 0.924 & 1.241 \\ 
+ generated pairs for ranking                       &  22,140 & \bf 0.347 & \bf 0.320 & 0.936 & 1.261 \\ 
+ generated data based on training human references &  59,436 & 0.343 & 0.278 & 0.922 & 1.238 \\ 
+ generated pairs for ranking                       & 163,764 & \bf 0.369 & 0.295 & 0.925 & 1.250 \\\hdashline[0.5pt/2pt]  
+ generated data based on test human references*    &  85,441 & 0.344 & 0.265 & 0.925 & 1.249 \\ 
+ generated pairs for ranking*                      & 236,578 & 0.345 & 0.256 & 0.944 & 1.277 \\\hline 
\end{tabular}
\caption{Results on the NEM ratings dataset; the number of training instances includes synthetic data (cf.\ Sections~\ref{sec:synth-data} and~\ref{sec:data}, Figure~\ref{fig:synth-pairs}). Boldface denotes configurations of our system that are significantly better than all previous ones according to the \citet{williams_regression_1959} test ($p<0.01$). Values for baseline metrics and the original RatPred system are taken over from \citet{dusek_referenceless_2017}. Configurations marked with ``*'' use human references for test instances (this includes word-overlap-based metrics such as BLEU).}
\label{tab:results-nem}
\end{table*}

\begin{table*}[tb]
\centering\small
\begin{tabular}{lrcc}\hline
\bf System & \bf Training insts & \bf Accuracy & \bf Avg.~loss \\\hline
Our base system & 11,921 & 0.708 & 0.173 \\  
+ generated pairs based on training system outputs & 50,324 & \bf \phantom{$^\dagger$}0.732$^\dagger$ & 0.158 \\  
+ generated pairs based on training human references & 428,873 & \bf 0.740 & 0.153 \\\hline 
\end{tabular}
\caption{Results on the E2E rankings dataset. Boldface denotes significant improvements over previous configurations according to pairwise bootstrap resampling \cite{koehn_statistical_2004} ($p<0.05$; $^\dagger=p<0.01$).}
\label{tab:results-e2e}
\end{table*}

The results on the NEM dataset in Table~\ref{tab:results-nem} 
show that our improved synthetic data generation methods bring significant improvements in correlation.\footnote{We used the \citet{williams_regression_1959} test to assess significant differences in correlation, following \citet{graham_testing_2014} and \citet{kilickaya_re-evaluating_2017}.} 
On the other hand, they worsen MAE and RMSE scores slightly, probably due to missing supervision on the exact rating in synthetic ranking instances.
Compared to \citet{dusek_referenceless_2017}, we get a 12\% increase in Pearson correlation on the best referenceless  configurations; our best referenceless system method outperforms even \citet{dusek_referenceless_2017}'s reference-aided system.
Note that the absolute correlations, while still not ideal, are much higher than those achieved by word-overlap-based metrics such as BLEU, which stay well below 0.1.

Our reference-aided setup did not improve with synthetic ranking pairs. 
Probably this is because there are already enough training data  for the domain. 
Furthermore, this system is more prone to overfit the validation set (exploiting validation references during training).
The intra-class correlation coefficient (ICC) of 0.45 measuring rater agreement on the NEM data as reported by \citet{novikova_why_2017} (`moderate agreement') also suggests that a certain level of noise may hinder further improvements on this dataset.


Table~\ref{tab:results-e2e} shows our results on the E2E data. Here, all configurations perform well above random chance (i.e.\ accuracy of 0.5). Using the synthesised ranking pairs brings a small but statistically significant\footnote{We used pairwise bootstrap resampling \cite{koehn_statistical_2004} to assess significance in ranking accuracy.} improvement over the base model (3\% using only NLG system outputs, additional 1\% if also human references from NLG system training data are used for synthetic pairs generation).


We also explored training the system using data from both sets; however, this did not bring performance improvements, probably due to different text styles in the two datasets -- the NEM data include requests, confirmations, etc., while the E2E data only contain informative statements.


%
%
\section{Related Work}\label{sec:related}

QE has been an active topic in many NLP tasks---image captioning \cite{anderson2016spice}, dialogue response generation \cite{lowe_towards_2017}, grammar correction \cite{napoles_theres_2016} or text simplification \cite{martin_reference-less_2018}---with MT being perhaps the most prominent area
\cite{specia_machine_2010,avramidis2012comparative,specia_quality_2018}.
QE for NLG recently 
saw an increase of focus in various subtasks, such as title generation \cite{ueffing_quality_2018,camargo_de_souza_generating_2018} or content selection and ordering \cite{wiseman2017challenges}. 
Furthermore, several recent studies focus on predicting NLG fluency only, e.g., \cite{tian_treat_2018,kann_sentence-level_2018}.

However, 
apart from our work,
\cite{dusek_referenceless_2017} is the only general NLG QE system to our knowledge,
which aims to predict the overall quality of a generated utterance, where quality includes both fluency and semantic coverage of the MR. Note that the correct semantic coverage of MRs is a problem for many neural NLG approaches \cite{gehrmann_end--end_2018,dusek_evaluating_2019,nie_simple_2019}. 
Compared to \citet{dusek_referenceless_2017}, our model is able to jointly rate and rank NLG outputs and includes better synthetic training data creation methods. 

Our approach to QE is similar to 
adversarial evaluation---distinguishing between
human- and machine-generated outputs 
\citep{goodfellow2014generative}. 
This approach is employed in generators for random text \cite{bowman2015generating} and dialogue responses \cite{kannan2017adversarial,li2017adversarial,bruni_adversarial_2017}. We argue that our approach is more explainable with users being able to reason with the ordinal output score.

\section*{Acknowledgments}

This research received funding from the EPSRC projects  DILiGENt (EP/M005429/1) and  MaDrIgAL (EP/N017536/1) and Charles University project PRIMUS/19/SCI/10.

\bibliographystyle{acl_natbib}
\bibliography{references}

\clearpage
\appendix
\onecolumn
\section*{Automatic QE for NLG: Ranting (Joint Rating and Ranking) -- Supplementary Material}

\begin{table}[H]
\centering
\begin{tabular}{l>{\centering\arraybackslash}m{3cm}}\hline
\bf Attribute & \bf Delexicalisation \\\hline
address & full \\
area & names only (excl.~\emph{city centre}, \emph{riverside})\\
customer rating & -\\
food/cuisine & -\\
kids-friendly & -\\
meal type & - \\
nearby venue/monument name\hspace{-1cm} & full\\
phone number & full\\
postcode & full\\
price & full\\
price range & -\\
venue count & full\\
venue name & full\\
venue type & - \\\hline
\end{tabular}
\caption{List of attributes in our data and their delexicalisation level. Delexicalised attributes are replaced with placeholders (e.g.\ ``X-name'' for venue names) to prevent data sparsity \cite{mairesse_phrase-based_2010,wen_semantically_2015}.}
\label{tab:delex}
\end{table}

\begin{table}[H]
\centering
\begin{tabular}{lc}\hline
\bf Parameter & \bf Value \\\hline
Embedding/GRU cell width & 50\\
Dropout keep rate & 0.8\\
Batch size & 50 \\
Learning rate & 0.0001 \\
Number of fully connected layers & 1\\
Max. number of epochs & 100\\
Epochs using synthetic data & 50\\\hline
\end{tabular}
\caption{Listing of network parameters used for our experiments on both datasets. Only the first two cross-validation folds from the NEM set and only the first 2,500 training instances from the E2E set were used for parameter search on development data.}
\label{tab:parameters}
\end{table}

\end{document}